# CNN based texture synthesize with Semantic segment

Xianye Liang, Bocheng Zhuo, Peijie Li, Liangju He

**Abstract:** Deep learning algorithm display powerful ability in Computer Vision area, in recent year, the CNN has been applied to solve problems in the subarea of Image-generating, which has been widely applied in areas such as photo editing, image design, computer animation, real-time rendering for large scale of scenes and for visual effects in movies. However in the texture synthesize procedure. The state-of-art CNN can not capture the spatial location of texture in image, lead to significant distortion after texture synthesize, we propose a new way to generating-image by adding the semantic segment step with deep learning algorithm as Pre-Processing and analyze the outcome.

**Key words:** convolutional neural network; semantic segment; texture synthesize

## 1. Introduction

Convolutional Neural Networks was first applied in document recognition and Le-net 5 achieve much higher recognition precision than other recognition algorithm [1]. The development of deep learning is quite exciting in recent years, particular in computer vision area, Over the past three years DCNNs have pushed the performance of computer vision systems to soaring heights on a broad array of high-level problems, including image classification(Krizhevsky[2] et al., 2013; Sermanet[3] et al., 2013; Simonyan & Zisserman[4], 2014; Szegedy[5] et al., 2014; Papandreou et al., 2014), fine-grained categorization (Zhanget[6] al., 2014), among others.( Semantic Image Segmentation with Deep Convolutional Nets and Fully Connected CRFs)

The way CNN model processing in image detection and recognition is simulating human being.in recent research, The features generate within CNN model is quite suitable for image-synthesis.Image-generating is mainly aim to make a new graph combine by content of an origin image and style(texture) of a target image, in general, there are two ways to texture synthesis.

**a.Procedural texture synthesis.** This method utilized biologically motivated method to generate texture. Greg Turk [7] texture an image by simulating a reaction-diffusion process on it. Walter and Fournier [8]  synthesis mammalian coat patterns based on cell division and cell-to-cell interactions. Although these method performs well the parameter of model is hard to optimized into ideal texture.

**b.Texture synthesis from samples.** In this method the texture is produce by one or some of target sample image, Zhu et. al. [9] model texture as a Markov Random Field and use Gibbs sampling for synthesis. Portilla 和 Simoncelli [10] propose a model combine all sort of wavelet features and their coefficients, it was considered to be the state of art model in texture synthesis. Efros [11] convert an input image into a specific stylization, through extract patterns as texture patches from existing artworks and proved to be a success in synthesizing in painting. Recently, generative models based on deep neural networks have shown exciting new perspectives for image synthesis [12,13,14,15]. Deep architectures capture appearance variations in object classes beyond the abilities of pixel-level approaches.

In general image synthesize procedure, the whole target image will be synthesized by one texture image, it will cause severe distortion in the objects of the target image. In this paper we propose to add semantic segmentation step before synthesize.

## 2.Image segment

In general image synthesize procedure, the whole target image will be synthesized by one texture image, it will cause severe distortion in the objects of the target image. In this paper we propose to add semantic segmentation step before synthesize. Once we get the origin image, we extract the foreground from the background (person, dog, cat, etc.), attribute to the development of deep learning in image semantic segmentation, which has been one of the most active topics in the field of image understanding and computer vision for a long time. we can get the foreground image directly in pixel level.

some of the recently proposed approaches to this task are based on the fully convolutional network(FCN) [16] trained end-to-end, pixels-to-pixels which is efficient and at the same time has achieved the state-of-the-art performance. By reusing the computed feature maps for an image, FCN avoids redundant re-computation for classifying individual pixels in the image. FCN becomes the standard approach to dense prediction and methods were proposed to further improve this framework, e.g., the DeepLab [17], deconvNet[18]and CRF-RNN[8]. One key reason for the success of these methods is that they are based on rich features learned from the very large ImageNet [19] dataset, often in the form of a 16-layer VGGNet [20]. However, currently, there exist much improved models for image classification, e.g., the ResNet [21, 22].

In this paper we compare four models in semantic segmentation in VOC 2012 test in Tab 1. We found that the CRF-RNN model has the best performance in accuracy, so we finally choose the CRF-RNN as our segment tool.

| | VOC 2012 test | | | |
|---|---|---|---|---|
| | FCN-8s | DeepLab | DeconvNet | CRF-RNN |
| MEAN IU | 62.2 | 71.6 | 72.5 | 74.7 |

**Table 1.** comparison of four different segment model in MEAN IU accuracy.

The key idea of CRF inference for semantic labelling is to formulate the label assignment problem as a probabilistic inference problem that incorporates assumptions such as the label agreement between similar pixels. CRF inference is able to refine weak and coarse pixel-level label predictions to produce sharp boundaries and fine-grained segmentations. Therefore, intuitively, CRFs can be used to overcome the drawbacks in utilizing CNNs for pixel-level labeling tasks. Importantly, with our formulation, the whole deep network, which comprises a traditional CNN and an RNN for CRF inference, can be trained end-to-end utilizing the usual back-propagation algorithm

The model comprises a fully convolutional network stage (used the FCN-8s architecture of [15], which provides unary potentials to the CRF. This network is based on the VGG-16 network), which predicts pixel-level labels without considering structure, followed by a CRF-RNN stage, which performs CRF-based probabilistic graphical modelling for structured prediction. The complete system, therefore, unifies strengths of both CNNs and CRFs and is trainable end-to-end using the back-propagation algorithm. and the Stochastic Gradient Descent (SGD) procedure.

During training, a whole image (or many of them) can be used as the mini-batch and the error at each pixel output of the network can be computed using an appropriate loss function such as the softmax loss with respect to the ground truth.

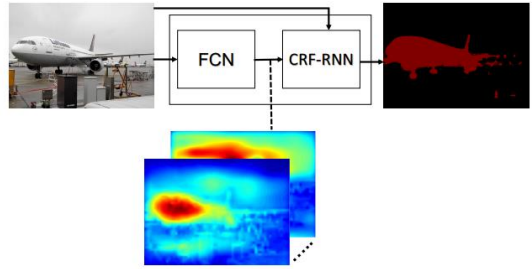

**Figure 1.** The End-to-end Trainable Network. Schematic visualization of the full network which consists of a CNN and the RNN-CRF network.

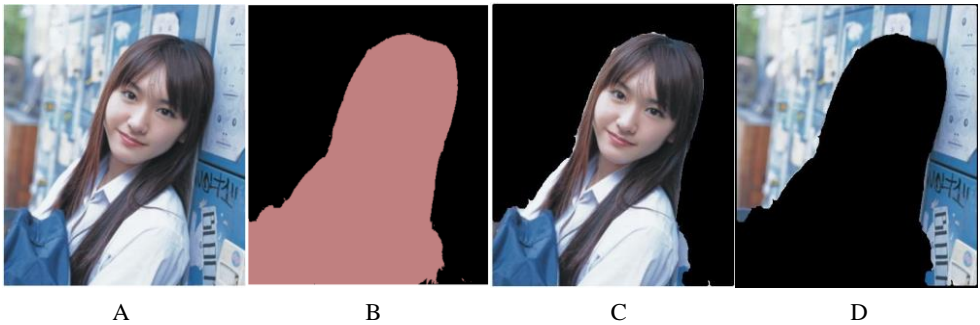

**Figure 2.** The procedure of semantic segment. (A)original image. (B) segment result. (C)extract foreground part. (D) extract background part.

## 3.Style matching

after extract the forward and backward image, we use the method in [13] to match the texture from style set. We compare AlexNet [2], GoogLeNet [5], andVGG-19, with AlexNet performing similarly to GoogLeNet and VGG-16 similarly to VGG-19. VGG networks perform much better at style transfer due to their architecture. For example, AlexNet and GoogLeNet strongly compress the input at the first convolutional layer using large kernels and stride (11 ×11 with stride 4 and 7 ×7 with stride 2 respectively) and thus a lot of fine detail is lost. VGG networks use 3 ×3 kernels with stride 1 at all convolutional layers and thus capture much more information. We have therefore used the VGG-19 network for our model.

To characterise a given vectorised texture $x$ in our model, we first pass $x$ through the convolutional neural network and compute the activations for each layer $l$ in the network. Since each layer in the network can be understood as a non-linear filter bank, its activations in response to an image form a set of filtered images (so-called *feature maps*). A layer with $N^l$ distinct filters has $N^l$ feature maps each of size $M_l$ when vectorised. These feature maps can be stored in a matrix $F^l \in R^{Nl \times Ml}$, where $F_{jk}^l$ is the activation of the $j^{th}$ filter at position $k$ in layer $l$. Textures are per definition stationary, so a texture model needs to be agnostic to spatial information. A summary statistic that discards the spatial information in the feature maps is given by the correlations between the responses of Textures are per definition stationary, so a texture model needs to be agnostic to spatial information. A summary statistic that discards the spatial information in the feature maps is given by the correlations between the responses of different features. These feature correlations are, up to a constant of proportionality, given by the Gram matrix $G^l \in R^{N_l \times N_l}$, where $G^l_{ij}$ is the inner product between feature map *i* and *j* in layer *l*:

$$G^l_{ij} = \sum_k F^l_{ik} F^l_{jk} \tag{1}$$

A set of Gram matrices $\{G^1, G^2, ..., G^L\}$ from some layers 1,2..., *L* in the network in response to a given texture provides a stationary description of the texture.

To generate a new texture on the basis of a given image, we use gradient descent from a white noise image to find another image that matches the Gram-matrix representation of the original image. This optimisation is done by minimised the mean-squared distance between the entries of the Gram matrix of the original image and the Gram matrix of the image being generated (Fig. 1B).

Let x and y be the original image and the image that is generated, and $G^l$ and $Q^l$ their respective Gram-matrix representations in layer *l* (Eq. 1). The contribution of layer *l* to the total loss is then

$$E_l = \frac{1}{4 N_l^2 M_l^2} \sum_{i,j} \left( G^l_{ij} - Q^l_{ij} \right)^2 \tag{2}$$

and the total loss is

$$Loss(x,y) = \sum_{l=0}^{L} w_l E_l \tag{3}$$

where $w_l$ are weighting factors of the contribution of each layer to the total loss. The derivative of $E_l$ with respect to the activations in layer *l* can be computed analytically:

$$\frac{\partial E_l}{\partial F^l_{ij}} = \begin{cases} \frac{1}{N_l^2 M_l^2} \left[ \left( F^l \right)^T \left( G^l_{ij} - Q^l_{ij} \right) \right]_{ji} & if \quad F^l_{ij} > 0 \\ 0 & if \quad F^l_{ij} < 0 \end{cases} \tag{4}$$

The gradients of *El*, and thus the gradient of *Loss*(*x,* y), with respect to the pixels *y* can be readily computed using standard error back-propagation [23]. The entire procedure relies mainly on the standard forward-backward pass that is used to train the convolutional network. Therefore, in spite of the large complexity of the model, texture generation can be done in reasonable time using GPUs and performance-optimised toolboxes for training deep neural networks [24].

in order to preserve the shape of forward image，we chose the forward image instead of white noise. we show textures generated by our model from different source images.as show in Fig3-B, the detail content been loss due to the spatial irrelevant in the feature map on synthesize procedure, by apply the segment procedure in target texture image Picasso, the detail content has been reserve in Fig3-C. The final synthesize result can be seen in Fig 4.

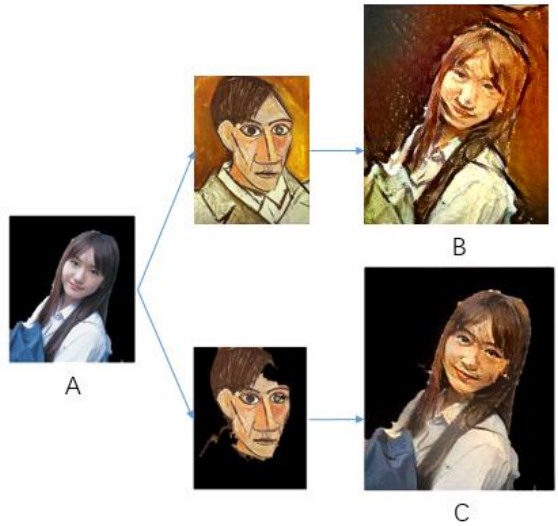

**Figure 3.** The procedure of style matching. (A)original image. (B) texture synthesize by the image Picasso without segment. (C) texture synthesize by the image Picasso with segment.

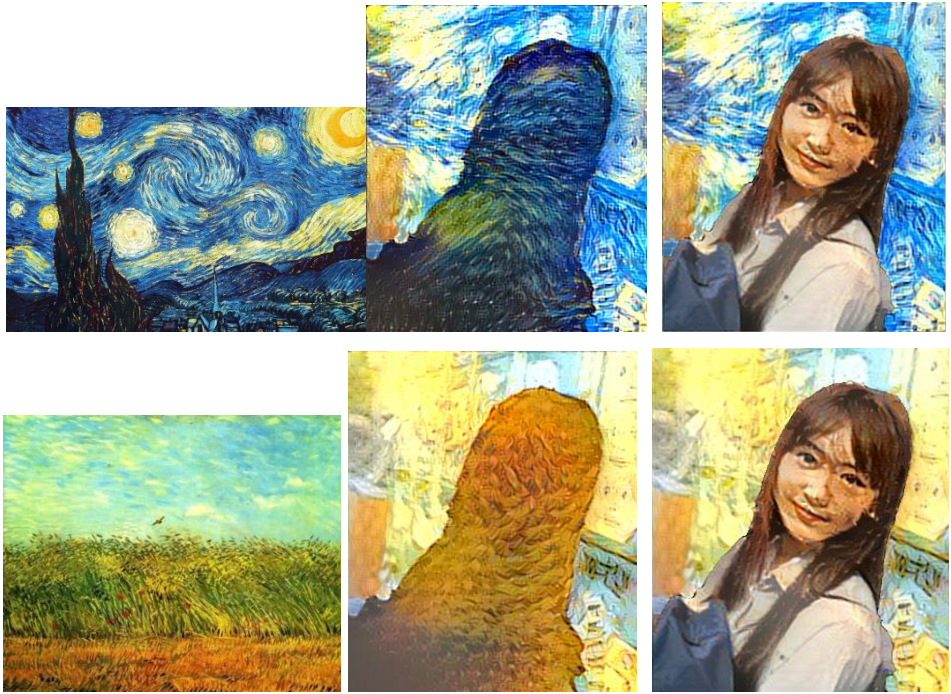

**Figure 4.** final result for texture synthesis

## 4.Conclusion

This paper presents a new way to texture synthesize which use two CNN(VGG-16 and VGG-19 respectively). This method reduced the texture distortion among different region in one image. In future research, establish an integrated texture database and intelligent texture matching algorithm will further realize the AI technology in painting.

1247–1283 (2000).